\documentclass[a4paper,conference]{IEEEtran}
\usepackage{cite}
\usepackage{amsmath,amssymb,amsfonts}
\usepackage{algorithmic}
\usepackage{graphicx}
\usepackage{textcomp}
\usepackage{xcolor}
\def\BibTeX{{\rm B\kern-.05em{\sc i\kern-.025em b}\kern-.08em
    T\kern-.1667em\lower.7ex\hbox{E}\kern-.125emX}}
\newcommand{\RN}[1]{%
  \textup{\uppercase\expandafter{\romannumeral#1}}%
}

\ifCLASSINFOpdf
\else
\fi
\begin{document}
%
\title{Learning Representation for Mixed Data Types with a Nonlinear Deep Encoder-Decoder Framework}

\author{\IEEEauthorblockN{Saswata Sahoo}
\IEEEauthorblockA{Gartner\\
Gurgaon, India\\
Email: saswata.sahoo@gartner.com}
\and
\IEEEauthorblockN{Souradip Chakraborty}
\IEEEauthorblockA{Walmart Labs\\
Bangalore, India\\
Email: souradip24@gmail.com}}


%


\maketitle

\begin{abstract}
Representation of data on mixed variables, numerical and categorical types to get suitable feature map is a challenging task as important information lies in a complex non-linear manifold. The feature transformation should be able to incorporate marginal information of the individual variables and complex cross-dependence structure among the mixed type of variables simultaneously.  In this work, we propose a novel nonlinear Deep Encoder-Decoder framework to capture the cross-domain information for mixed data types. The  hidden  layers of the network  connect  the  two types  of  variables  through  various  non-linear  transformations to give latent feature maps. We encode the information on the numerical variables in a number of hidden nonlinear units. We use these units to recreate categorical variables through further nonlinear transformations. A separate and similar network is developed switching the roles of the numerical and categorical variables. The hidden representational units are stacked one next to the others and transformed into a common space using a locality preserving projection. The derived feature maps are used to explore the clusters in the data. Various standard datasets are investigated to show nearly the state of the art performance in clustering using the feature maps with simple K-means clustering.
\end{abstract}


%
\IEEEpeerreviewmaketitle

\section{Introduction}
Data points on mixed type of variables, numerical and categorical types frequently appear in data science applications. Numerical variables take values on the real line within the range of variations whereas, categorical variables indicate class label information.
The joint distribution of mixed variables lies in a complex non-linear product space. It is important to represent the data points on a common feature space using suitable feature transformation so that maximum information on the marginal and joint distributions are incorporated in the representation. Representation of the data points can be carried out in a supervised or an unsupervised manner. In a supervised framework, there is some dependent variable associated to the data points. Feature transformation in such a case is essentially finding a suitable transformation of the mixed type of variables which optimises the predictability of the dependent variable. It is quite evident that such transformations are heavily dependent on the nature of the prediction problem. In this work, we are rather interested in the unsupervised framework, where we have no dependent study variable or any ground truth class label information on the data points. The study focuses on exploring a general feature transformation strategy for mixed type of data points in an unsupervised setup. It is important to ensure that the transformed feature map retains strong signals present in the actual data distribution and preserve the original distance geometry. We propose a framework based on a Deep encoder-decoder network and a locality preserving transformation to represent the mixed data points by a dense feature set. 
\begin{figure*}[ht]
  \includegraphics[width=\textwidth,height=8cm]{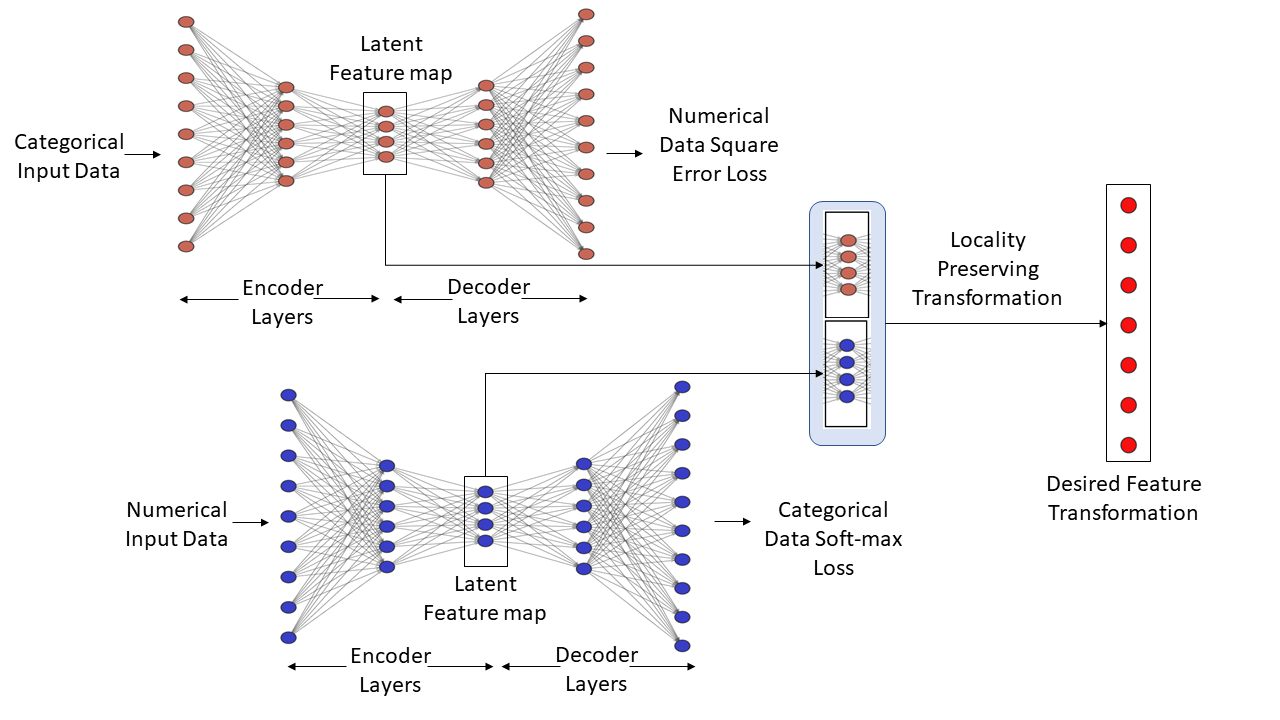}
  \caption{The complete network architecture}
  \label{arch}
\end{figure*}

\section{Previous Work}
There are various investigations by researchers to represent data points in unsupervised framework. Linear projection on the  Principal component space introduced by works well when majority of the variations of the data points lie in a linear subspace (see \cite{pca,spca} ). In absence of such linear subspace, data points are transformed to higher dimensional space using nonlinear Kernel feature maps (see \cite{kpca}).
Fourier feature map involves sampling components from frequency space of the Kernel functions and embedding the data points using the random feature maps (see \cite{fourier}). There are some localised approaches involving locality based manifold learning such as Local linear embedding \cite{lle} or Isometric feature mapping \cite{isomap}. Multidimensional scaling gives feature representation \cite{mds} preserving mutual dissimilarity. On the other hand, a feature representation is given by embedding the data points on the eigen space of the graph Laplacian of the affinity graph (see \cite{graph}). Deep Learning based architecture combines the variables using various nonlinear functions in iterative fashion to incorporate different degrees of non-linearity (see ~\cite{dl}). Deep Kernel networks are used to propagate information from one layer to the next layer for a sequentially feature maps of the data points using suitable kernel functions (see\cite{kernel}).
\par
For mixed data types, one straight forward approach involves converting numerical variables to categorical variables through discretization (see \cite{discretize}). On the other hand, the categorical variables can be converted to numerical variables through dense embedding utilising the intrinsic low rank structure.(see \cite{catembed1},\cite{catembed2}, \cite{catembed3}).  Another class of techniques involve considering pairwise dissimilarities of the two type of data separately. \cite{kononenko} suggested using Euclidean distance for numerical variables and the Hamming distance for categorical variables. Combining the two types of distances, a latent feature map is recovered. \cite{tang} proposed a feature selection for mixed data type, based on mixed feature subset evaluation. The techniques involve first decomposing the feature space based on categorical variables and then measuring class separability based on numerical features in each of the subspaces. \cite{sahoo} considered the mixed variables as independent random samples on the nodes of an undirected graph. The edges of the graph are estimated using a suitable mapping function encoding the mutual dependence of the nodes. The desired feature transformation reduces to embedding the mixed data points on the eigen space of the Graph Laplacian.

\section{Present Work}
 In this work we propose a framework to a define a  feature transformation which connects the two types of data space. We place the numerical variables in one side and the categorical variables in the other side of a deep encoder-decoder network. We encode the categorical variables into a set of nonlinear latent feature maps. The feature maps are further sequentially combined  through a set of layers of nonlinear functions to recreate the numerical variables. The  parameters of the network of functions are optimized to preserve maximum information of the categorical variables and maximize the accuracy of recreating the numerical variables. The latent feature maps are treated as foundational units derived from categorical data and can be combined to recreate the numerical data. So the units can be thought of preserving information from both the spaces with different degrees.  We create a separate set of latent feature maps switching the roles of numerical and categorical variables in a very similar second network. The concatenated feature maps from the two networks are now  treated as a set of units having higher order nonlinear information and regenerating abilities borrowed from the two spaces. The concatenated feature maps are projected using a locality preserving transformation to get the final feature map of the mixed data types.
\par
The core of the strategy is to embed the mixed type of data to a common latent feature space to ensure optimal information flow from the two spaces. It also involves preserving the local information of the data points in the transformed feature map. This strategy is quite a significant improvement over the past technique of handling mixed data types. The proposed feature map shows directional signs of preserving important information of the mixed data points by showing excellent performance in naive K-means clustering. The feature maps with naive K-means clustering  performs equally well, and in cases, better than some of the complex clustering techniques available in literature for mixed data types. The key points we try to make here are as follows: 
\begin{enumerate}
    \item Complex nonlinear structure of mixed type variables confounded in the mixed data product space is recoverable through well designed nonlinear transformations.
    \item Nonlinear projections of mixed type of data points to a common feature space can retain information from both the spaces.
    \item  With high quality dense nonlinear feature maps even a simple clustering algorithm such as k-means, can perform equally well as some of the complex competitor clustering techniques for mixed data types.
\end{enumerate}

\par
The rest of the paper is organized as follows. In {\bf section \RN{4}} we describe the general framework and notations. {\bf Section \RN{5}} is dedicated to describing the proposed methodology in details.  Clustering is one of most basic tasks of data mining. In {\bf Section \RN{6}}  we describe the effect of the feature transformation on the performance of  clustering. We conclude the paper with a brief discussion in {\bf section \RN{7}}
\begin{figure*}[ht]
\centering
  \includegraphics[width=\textwidth,height=8cm]{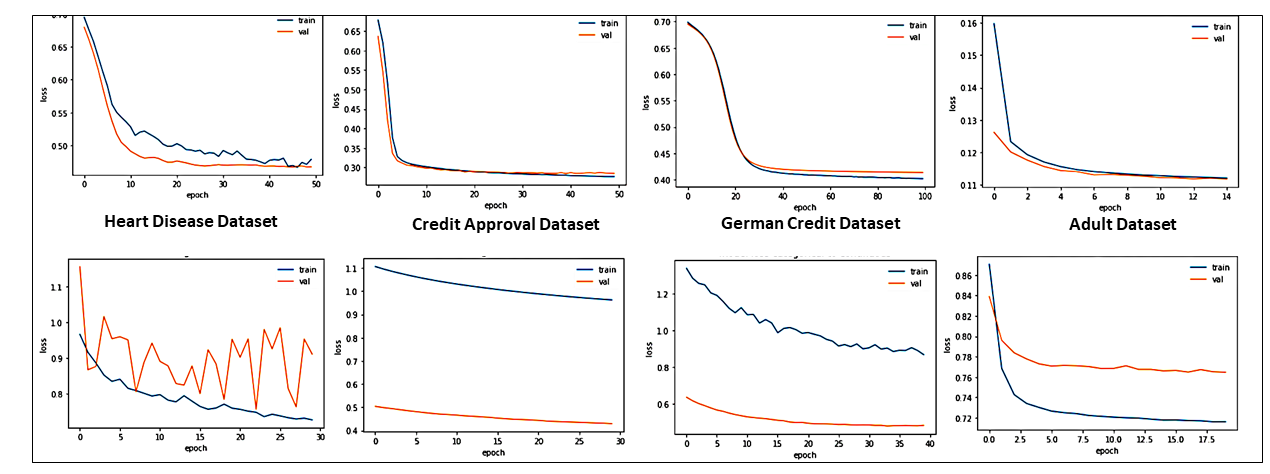}
  \caption{The values of the soft-max binary cross entropy loss and means square error loss function for the different epochs of estimating the deep hidden layer networks are give for the 4 datasets. The plots at the top indicate binary cross entropy  loss when the input is numerical data and output is the categorical data. The plots at the bottom indicate mean square error  loss for the network where the input is categorical data and output is the numerical data. The blue curve denotes the training loss and the yellow curve denotes the validation loss.  }
  \label{loss}
\end{figure*}
\section{General Framework}
Let us consider $p_1$ numerical variables denoted by $X^{(num)'}=(X^{(num)}_1, X^{(num)}_2,\ldots, X^{(num)}_{p_1})$, assuming values in $p_1$ dimensional Euclidean space. There are $l$ categories denoted by  $C_1,C_2,\ldots, C_l$. The categories have multiple levels. However, for each of the $l$ categories, different levels can be represented by a binary valued dummy variable. For example, for the $j$th category, let us assume there are $m_j$ distinct category levels, denoted by $C_j(1), C_j(2),\ldots, C_j(m_j).$ We define a dummy indicator variable $Z_{C_j(u)}$ assuming values 1 or 0, according as the level $C_j(u)$ is present or not, for $u=1,2,\ldots, m_j$, $j=1,2,\ldots, l$. For each of the $l$ categories, we define similar dummy indicator variables and arrange all of them together in a random vector  $X^{(cat)}$. This random vector has $p_2$ binary valued random variables,  where $p_2=\sum_{j=1}^lm_j$. We denote the mixed type of variables by  $X=(X^{(num)'},X^{(cat)'} )'$ of dimension $p=p_1+p_2$.  
There are $n$ independent $p$ dimensional observations on the random vector $X$ denoted by $x_{(i)}=(x_{(i)1},x_{(i)1}, \ldots, x_{(i)p})'$, $i=1,2,\ldots,n$. The complete $n\times p$ dimensional data matrix of the $n$ observations is denoted by $\mathcal{D}=[x_{(1)}, x_{(2)},\ldots, x_{(n)}]'$, the rows being the independent observations on the random vector $X$. The data matrix $\mathcal{D}$ is partitioned into two submatrices as $\mathcal{D}=[\mathcal{D}^{(num)},\mathcal{D}^{(cat)}]$. The submatrix $\mathcal{D}^{(num)}$ is of dimension $n\times p_1$, consisting of the $n$ rows of the numerical observations on the $p_1$ dimensional numerical random vector $X^{(num)}$ and the submatrix $\mathcal{D}^{(cat)}$ is of dimension $n\times p_2$, consisting of the $n$ rows of the categorical observations on the $p_2$ dimensional categorical random vector $X^{(cat)}.$ The main problem is to come up with a feature transformation $\phi(.)$ which maps the $p$ dimensional mixed observation $x_{(i)}$ to $\phi(x_{(i)})$ giving the desired feature map of mixed data type. 

\section{Proposed Methodology}
In this section we describe the proposed framework in details. Consider Fig \ref{arch} for the complete network architecture. First, we consider a network of encoder layers which takes the numerical data points as input and creates a latent feature map. The latent feature map is non-linearly combined through multiple layers to finally connect to a soft-max function. The soft-max function is optimized to recover class label information of the categorical variables. We create a similar network, where the input is now the categorical variables  and the final output is connected to mean square error loss function of the numerical variables. The latent feature maps from the two networks are concatenated and projected on a space which preserves the localised information of the data points. The full architecture gives us a feature transformation for the mixed type of data points as final output.
\subsection{Latent Feature Map}
Let us consider the $i$th row of the numerical variable data matrix $\mathcal{D}^{(num)}$ denoted by $x^{(num)}_{(i)}$ and for that of the categorical variable data matrix $\mathcal{D}^{(cat)}$ denoted by $x^{(cat)}_{(i)}$. We take $x^{(num)}_{(i)}$ as the input data and create a latent feature map using an encoder network consisting of  $\kappa_1$ layers. 
We denote the $\kappa_1$ encoder layers by $\mathcal{E}_v$, $v=1,2,\ldots, \kappa_1$ where the $v$th encoder layer is a mapping  $\mathcal{E}_v:\mathbb{R}^{d_{v-1}}\rightarrow \mathbb{R}^{d_{v}}$ with $p_1=d_{0}>d_1>\ldots>d_{\kappa_1}.$
For the $i$th observation $x^{(num)}_{(i)}\in \mathbb{R}^{p_1}$ the $v$th layer of the encoder network $\mathcal{E}_v$  has input from $d_{v-1}$ channels denoted by
\[Z^{v-1}_{(i)}=(Z^{v-1}_{(i),1},Z^{v-1}_{(i),2},\ldots,Z^{v-1}_{(i),d_{v-1}})\in \mathbb{R}^{d_{v-1}}\] and it maps the input to  \[Z^{v}_{(i)}=(Z^{v}_{(i),1},Z^{v}_{(i),2},\ldots,Z^{v}_{(i),d_{v}})\in \mathbb{R}^{d_{v}}.\] The components of the output of the map are defined as
$Z^{v}_{(i),l}=\sigma(A^{(v)'}_lZ^{v-1}_{(i)}),$ where $A^{(v)}_l,$ for $l=1,2,\ldots, d_{v}$ are the layer parameter vectors and $\sigma(.)$ is the element-wise rectified linear unit activation function (ReLu). The complete encoder network consisting of $\kappa_1$ layers maps the input data $Z^{0}_{(i)}=x^{(num)}_{(i)}\in \mathbb{R}^{p_1}$ to a feature map 
\begin{equation}
    \centering
    Z^{\kappa_1}_{(i)}=y^{(num)}_{(i)}\in\mathbb{R}^{d_{\kappa_1}}. 
    \label{num_cat}
\end{equation}
\par
The latent feature map from the full encoder network,   $y^{(num)}_{(i)}\in\mathbb{R}^{d_{\kappa_1}}$ is sent through a set of decoder layers.  We keep similar choices of number of layers, neurons and activation function for the decoder network as it was for the encoder network. The decoder layers are denoted by $\overline{\mathcal{E}}_v:\mathbb{R}^{d_{v}}\rightarrow \mathbb{R}^{d_{v-1}}$ for $v=2,\ldots, \kappa_1$, in the reverse order: $\overline{\mathcal{E}}_{\kappa_1},\overline{\mathcal{E}}_{\kappa_1-1},\ldots, \overline{\mathcal{E}}_{2} $ with $d_{\kappa_1}<d_{\kappa_1-1}<\ldots<d_{1}.$
The final decoder layer $\overline{\mathcal{E}}_1:\mathbb{R}^{d_{1}}\rightarrow \mathbb{R}^{p_2}$ gives an output of the form 
 \[\overline{Z}^{1}_{(i)}=(\overline{Z}^{1}_{(i),1},\overline{Z}^{1}_{(i),2},\ldots,\overline{Z}^{1}_{(i),{p_2}})\in \mathbb{R}^{p_2}.\]
$\overline{Z}^{1}_{(i)}$ is connected to a soft-max layer giving the probability of the categorical variable class levels. For the $i$th observation on the categorical variable $x^{(cat)}_{(i)}$, the corresponding soft-max function takes the form:
\begin{equation}
  \centering
  \prod_{l=1}^{p_2}\bigg[\frac{exp(\overline{Z}^{1}_{(i),l})}{\sum_{l'=1}^{p_2}exp(\overline{Z}^{1}_{(i),l'})}\bigg]^{1(x^{(cat)}_{(i)l}=1)}
  \label{softmax}
\end{equation}
where $1(x^{(cat)}_{(i)l}=1)$ denotes an indicator variable taking value 1 or 0 according as $x^{(cat)}_{(i)l}=1$ or $x^{(cat)}_{(i)l}=0$.
\begin{table}[t]
\centering
\caption{Choice of the number of hidden layers $(\kappa_1, \kappa_2)$ and eigen vectors ($L$) to optimize RI and NMI.  }
\label{selection_of_hyperparameters}
\begin{tabular}{|c|ccc|c|c|} 
\hline
Dataset                 & $\kappa_1$ & $\kappa_2$ & L~ & RI     & NMI     \\ 
\hline
Heart Disease Dataset   & 5           & 7                            & 30 & 0.7162 & 0.3454  \\
Credit Approval Dataset & 3           & 5                            & 79 & 0.7034 & 0.3389  \\
German Credit Dataset   & 4           & 4                            & 14 & 0.5501 & 0.0218  \\
Adult Dataset           & 5           & 6                            & 90 & 0.6202 & 0.0924   \\
\hline
\end{tabular}
\end{table}
\par
We define a similar network switching the roles of the numerical and categorical variables. The $i$th observation on the categorical variable $x^{(cat)}_{(i)}\in \mathbb{R}^{p_2}$ is sent through an encoder network consisting of $\kappa_2$ layers.  The complete encoder network maps the input data $x^{(cat)}_{(i)}\in \mathbb{R}^{p_1}$ to a feature map 
\begin{equation}
    \centering
    y^{(cat)}_{(i)}\in\mathbb{R}^{d_{\kappa_2}}.
\label{cat_num}
\end{equation}

The latent feature map from the full encoder network,   $y^{(cat)}_{(i)}\in\mathbb{R}^{d_{\kappa_2}}$ is sent through a set of decoder layers. The final decoder layer  gives an output of the form 
 \[\overline{Z^*}^{1}_{(i)}=(\overline{Z^*}^{0}_{(i),1},\overline{Z}^{*0}_{(i),2},\ldots,\overline{Z^*}^{0}_{(i),{p_1}})\in \mathbb{R}^{p_1}.\]
$\overline{Z^*}^{1}_{(i)}$ is connected to mean square error function giving the error in estimating the numerical variables. For the $i$th observation on the numerical variable $x^{(num)}_{(i)}$, the corresponding mean-square error function takes the form:
\begin{equation}
    (p_1)^{-1}\left\Vert x^{(num)}_{(i)}-\overline{Z^*}^{1}_{(i)}\right\Vert^2,
    \label{mse}
\end{equation} where $\left\Vert.\right\Vert$ is the $L2$ norm. 
\par
From the two networks, the encoder layer latent feature maps given in \eqref{num_cat} and \eqref{cat_num} are concatenated together to form a latent embedding of the mixed variables of the form:
\begin{equation}
    \centering
    w_{(i)}=[y^{(cat)'}_{(i)},y^{(num)'}_{(i)}]'\in \mathbb{R}^{d_{\kappa_1}+d_{\kappa_2}}, i=1,2,\ldots, n.
    \label{concat}
\end{equation}
\begin{table*}[t]
\centering\caption{The clustering results with respect to RI and NMI for varying choices of number of dimensions of the final feature map (L) for different datasets. The best choices of dimensions are given in bold font.}
\label{vary_ev}
\begin{tabular}{|c|cc|c|cc|c|cc|c|cc|} 
\hline
\multicolumn{3}{|c|}{Heart Disease Dataset} & \multicolumn{3}{c|}{Credit Approval Dataset} & \multicolumn{3}{c|}{German Credit Dataset} & \multicolumn{3}{c|}{Adult Dataset} \\ 
\hline
L & RI & NMI & L & RI & NMI & L & RI & NMI & L & RI & NMI \\ 
\hline
5 & 0.7110 & 0.3370 & 20 & 0.6640 & 0.2620 & 5 & 0.5424 & 0.0070 & 40 & 0.5324 & 0.0245 \\
15 & 0.7065 & 0.3257 & 30 & 0.6730 & 0.2720 & 10 & 0.5500 & 0.0130 & 60 & 0.6145 & 0.0866 \\
\textbf{30} & \textbf{0.7162} & \textbf{0.3454} & 50 & 0.6800 & 0.2840 & \textbf{14} & \textbf{0.5500} & \textbf{0.0220} & 80 & 0.6174 & 0.0893 \\
45 & 0.7157 & 0.3450 & \textbf{79} & \textbf{0.7034} & \textbf{0.3389} & 20 & 0.5474 & 0.0193 & \textbf{90} & \textbf{0.6202} & \textbf{0.0924} \\
60 & 0.7070 & 0.3384 & 100 & 0.6950 & 0.3316 & 30 & 0.5030 & 0.0004 & 120 & 0.6152 & 0.0884 \\
90 & 0.7002 & 0.3374 & 120 & 0.6950 & 0.3316 & 50 & 0.5010 & 0.0002 & 140 & 0.6135 & 0.0867 \\
120 & 0.6989 & 0.3301 & 140 & 0.6820 & 0.3149 & 60 & 0.5020 & 0.0001 & 160 & 0.6112 & 0.0846 \\
\hline
\end{tabular}
\end{table*}
\subsection{Locality Preserving Transformation}
The concatenated latent feature maps from the two networks, denoted by $w_{(i)}=[y^{(cat)'}_{(i)},y^{(num)'}_{(i)}]'$, for $i=1,2,\ldots,n$ are suitably transformed to preserve the local geometry of the actual data points $\{x_{(i)}, i=1,2,\ldots, n\}$. To summarise the information of the local geometry of the data points we consider a suitable kernel function $s(.,.)$.
A kernel matrix $S\in \mathbb{R}^{n\times n}$ for the data points  $x_{(i)}, i=1,2,\ldots,n$ is constructed using the chosen kernel function. 
The kernel matrix contains the information on the localised structure of the data points in the original mixed data space. We use the vector space spanned by the kernel matrix to define the desired locality preserving transformation. We propose to estimate a transformation matrix $V\in\mathbb{R}^{(d_{\kappa_1}+d_{\kappa_2})\times L}$ so that the latent feature embedding obtained in \eqref{concat} can be suitably transformed to an $L$ dimensional space as $\phi(x_{(i)})=V'w_{(i)}$ preserving the localised information in the Kernel matrix. The transformation matrix $V$ is estimated by minimizing the objective the function 
\begin{equation}
\sum_{i,i'=1}^{n}s_{ij}\left\Vert V'w_{(i)}-V'w_{(j)}\right\Vert^2,
\label{3rdobj}    
\end{equation}
where $s_{ij}$ is the $(i,j)$th element of the kernel matrix $S$.
We put a penalty for mapping nearest neighbouring points in the actual space to distant points after the projection. So, for pair of points in the actual data space $(x_{(i)},x_{(j)})$, the distance between the projected data points 
$(V'w_{(i)},V'w_{(j)})$ are penalized by the kernel value $s_{ij}.$
\par
Denote the degree-matrix of $S$ by the diagonal matrix $\Lambda$ with diagonal entries being  $\lambda_{ii}=\sum_{i'=1}^n s_{ii'}.$
This optimization in \eqref{3rdobj} is identical to the following problem:
\begin{equation}
    \min_{\substack{V\in\mathbb{R}^{(d_{\kappa_1}+d_{\kappa_2})\times L}\\V'W\Lambda W'V=I }} Tr[V'W(\Lambda-S)W'V]
\label{locality}
\end{equation}
\subsection{Full Network Optimization}
Combining the two networks and the  locality preserving transformation, the full objective function is given by 
\begin{equation}
    \begin{split}
        \min_{\substack{\Theta_{num}, \Theta_{cat}\\
        V\in\mathbb{R}^{(d_{\kappa_1}+d_{\kappa_2})\times L}\\V'W\Lambda W'V=I \\ }}\alpha J_{1}(\Theta_{num})+(1-\alpha)J_{2}(\Theta_{cat})+\beta P(V).
    \end{split}
    \label{full}
\end{equation}
The first term $J_{1}(\Theta_{num})$ corresponds to the loss function associated to the soft-max function given in \eqref{softmax}:
\[J_{1}(\Theta_{num})=-log\bigg[\prod_{i=1}^n\prod_{l=1}^{p_2}\bigg[\frac{exp(\overline{Z}^{0}_{(i),l})}{\sum_{l'=1}^{p_2}exp(\overline{Z}^{0}_{(i),l'})}\bigg]^{1(x^{(cat)}_{(i)l}=1)}\bigg].\] The second term corresponds to the 
mean square error loss function given in \eqref{mse}
denoted by
\[J_2(\Theta_{cat})= (np_1)^{-1}\sum_{i=1}^n\left\Vert x^{(num)}_{(i)}-\overline{Z^*}^{1}_{(i)}\right\Vert^2.\]
The third term is treated as a penalty function incorporating the locality preserving transformation given by 
\[P(V)=\sum_{i,i'=1}^{n}s_{ij}\left\Vert V'w_{(i)}-V'w_{(j)}\right\Vert^2.\] The parameters involved in the two proposed deep encoder decoder networks are given by $\Theta_{num}$ and $ \Theta_{cat}$. We introduce a tuning parameter for the penalty function denoted by $\beta>0$ which controls how much the linear transformation $w_{i}\mapsto V'w_{i}$ preserves the feature map from the networks and how much it retains localised structure of the data points $\{x_{(i)}, i=1,2\ldots, n\}.$ The parameter $\alpha\in (0,1)$ is used to combine the loss functions from the two networks and controls the flow of the relative information from the two networks to the final feature map $\phi(x_{(i)}).$
\par
We estimate the network parameters using gradient decent methods on the objective function $J_{1}(\Theta_{num})+(1-\alpha)J_{2}(\Theta_{cat})$. The estimated latent feature maps denoted by $w_{(i)}=[y^{(cat)'}_{(i)},y^{(num)'}_{(i)}]'$ for $i=1,2,\ldots, n$ are used to solve the generalized eigen value problem corresponding to the locality preserving penalty function $P(V)$ to estimate V:
\begin{equation}
W(\Lambda-S)W'v_j=\eta_jW\Lambda W'v_j
\label{gen}
\end{equation} 
where the eigen values are given by $\eta_j, j=1,2,\ldots, L$ and the eigen vectors are $v_j,j=1,2,\ldots,L$. The above problem is solved and  by taking the estimated eigen vectors the transformation matrix $V$ is constructed. The penalization function is calculated plugging its value in $P(V)$. The complete objective function in \eqref{full} is optimized till convergence and the final feature map is estimated as $x_{(i)}\mapsto\phi(x_{(i)})=V'w_{(i)}.$

\section{Numerical Investigation}
In this section, we numerically investigate the quality of the proposed feature transformation based on the performance of data clustering. Our hypothesis is that, a simple K-means clustering implemented on the derived feature maps should perform equally well, if not outperforming many sophisticated clustering algorithms applied on the original data points.

\subsection{Datasets}
We consider 4 publicly available datasets from \cite{datasets} consisting of mixed type of variables to conduct the experiments. The {\it Heart Disease Dataset} consists of 303 observations related to diagnosis of heart disease. The dataset has 7 categorical variables and 6 numerical variables. The categorical variables have multiple levels. We define multiple dummy binary variables to indicate different levels of the categorical variables. As a result, there are 20 mixed type of variables consisting of numerical and binary valued categorical variables.  The observations have two  partitions with ground truth class labels: healthy and sick. The {\it  Credit Approval Data Set} has 690 observations on credit card applications. The data has 9 categorical variables and 6 numerical variables describing features of the individual credit card applications. After redefining the categorical variables with binary dummy variables, there are 46 mixed types of variables in the dataset. There are two partitions with ground truth class labels on whether the credit card application is approved or not. The {\it German Credit Data} has information on 1000 individuals covering 13 categorical variables and 7 numerical variables describing various attributes of the individuals. The total number of mixed type of variables consisting of numerical and binary categorical variables is 71.  There are  partitions with ground truth labels for the different observations describing whether each individual has a good credit risk or bad credit risk. The 4th dataset named as {\it Adult Data Set } is quite large, consisting of 45222 data points and the data points are having  8 categorical variables  and 6 numerical variables. There are 108 mixed types of variables after redefining the multi-level categorical variables with binary dummy variables. The data points are grouped into two partitions.
 
\begin{table*}
\centering
\caption{The clustering results with respect to RI and NMI for varying choices of number of clusters for different datasets}
\label{vary_clusters}
\begin{tabular}{|c|cc|c|cc|c|cc|c|cc|} 
\hline
\multicolumn{3}{|c|}{Heart Disease Dataset} & \multicolumn{3}{c|}{Credit Approval Dataset} & \multicolumn{3}{c|}{German Credit Dataset} & \multicolumn{3}{c|}{Adult Dataset} \\ 
\hline
\begin{tabular}[c]{@{}c@{}}No. of \\Clusters \end{tabular} & RI & NMI & \begin{tabular}[c]{@{}c@{}}No. of \\Clusters \end{tabular} & RI & NMI & \begin{tabular}[c]{@{}c@{}}No. of \\Clusters \end{tabular} & RI & NMI & \begin{tabular}[c]{@{}c@{}}No. of \\Clusters \end{tabular} & RI & NMI \\ 
\hline
2 & 0.716 & 0.3454 & 2 & 0.7034 & 0.3389 & 2 & 0.5501 & 0.0218 & 2 & 0.6202 & 0.0924 \\
4 & 0.5890 & 0.3109 & 4 & 0.6150 & 0.3395 & 4 & 0.4713 & 0.0258 & 4 & 0.4735 & 0.1704 \\
6 & 0.5891 & 0.3670 & 6 & 0.5919 & 0.3762 & 6 & 0.4642 & 0.0367 & 6 & 0.4569 & 0.1715 \\
8 & 0.5614 & 0.4006 & 8 & 0.5718 & 0.3681 & 8 & 0.4513 & 0.0551 & 8 & 0.4277 & 0.1934 \\
10 & 0.5338 & 0.3612 & 10 & 0.5486 & 0.4120 & 10 & 0.4455 & 0.0567 & 10 & 0.4286 & 0.2188 \\
20 & 0.5230 & 0.3891 & 20 & 0.5287 & 0.4905 & 20 & 0.4327 & 0.0626 & 20 & 0.3956 & 0.2161 \\
30 & 0.5180 & 0.4248 & 30 & 0.5212 & 0.5211 & 30 & 0.4281 & 0.0734 & 30 & 0.3884 & 0.2290 \\
\hline
\end{tabular}
\end{table*}

\subsection{Performance Metrics}
 We judge the feature maps based on the quality of clusters they produce using simple K-means clustering algorithm. We compare the performance of the clusters with some state of the art competitor clustering techniques applied on the  data points on the actual mixed variable space.
 There are various performance metrics available in the literature on the quality of the clustering. We prefer the following two metrics: Rand index (RI) ,
and normalized mutual information (NMI) proposed by \cite{rand} and \cite{nmi} respectively. RI is defined as 
\begin{equation}
    RI=\frac{TP + TN}{TP + FP + FN + TN}
    \label{ri}
\end{equation}
where TP, TN, FP, and FN stand for true positive, true
negative, false positive, and false negative with respect to the ground truth class label information. The other metric NMI is defined as 
\begin{equation}
    NMI=\frac{\sum_{j=1}^k\sum_{j'=1}^{k^*}N_{jj'}log\frac{NN_{jj'}}{N_{j}N_{j'}}}{\big[(\sum_{j=1}^{k}N_j\log\frac{N_j}{N})(\sum_{j'=1}^{k^*}N_j\log\frac{N_{j'}}{N})\big]^{1/2}}.
    \label{nmi}
\end{equation}

Here $N$ is the total number of observations,  $k^*$ is the true number of clusters in the data and $k$ denotes the number of clusters obtained using the chosen algorithm. The number of agreements between the cluster $j$ and true class $j'$ is denoted by $N_{jj'}$. The number of observations in cluster $j$ and true class $j'$ are given by $N_{j}$ and $N_{j'}$ respectively. RI is a pair-counting measure and it calculates how the assigned cluster labels agree with the true partition labels. Whereas, NMI is a measure based on mutual information between the true data partition and obtained clusters. The values of both the metrics RI and NMI lie within the interval $[0, 1]$ and larger value of the metrics indicate improved performance of the clustering. 

\subsection{Choice of Network Hyper-parameters}
 There are various hyper parameters involved in the proposed feature transformation methodology. The performance of the proposed feature transformation can vary depending on the choice of the hyper parameters. 
 There are two separate networks in the proposed framework: The first networks has numerical data as the input and through a number of hidden layers the final output is a fully connected soft-max layer for the categorical data. Here $\kappa_1$ denotes number of hidden layers for this network. For the second network, categorical data are given as input and the numerical data output are connected to mean square error loss function.
Here $\kappa_2$ denotes the number of hidden layers for this network. The kernel function used to derive the locality preserving transformation is chosen to be polynomial kernel. The number of eigen vectors of the generalised eigen value problem given in \eqref{gen}  to construct the transformation matrix decides the dimension of the final feature map. The number of eigen vectors selected to get the transformation matrix is denoted by $L$.
\par

\begin{table*}
\centering
\caption{The clustering results with respect to RI and NMI for different standard clustering techniques as compared to the simple K-means clustering on the proposed feature maps for different datasets}
\label{compare_cluster}

\refstepcounter{table}
\label{comparison}
\begin{tabular}{|c|cccc|c|cccc|c|} 
\hline
Methods & WKM & EWKM & OCIL & WOCIL & \begin{tabular}[c]{@{}c@{}}\textbf{KMFM}\\\textbf{(Proposed }\\\textbf{Method)}\end{tabular} & WKM & EWKM & OCIL & WOCIL & \begin{tabular}[c]{@{}c@{}}\textbf{KMFM }\\\textbf{(Proposed }\\\textbf{Method)}\end{tabular} \\ 
\hline
{Metrics} & \multicolumn{4}{c|}{RI} & \textbf{RI }  & \multicolumn{4}{c|}{NMI} & \textbf{NMI }  \\ 
\hline
Heart Disease dataset & 0.6183 & 0.6431 & 0.6473 & 0.6894 & \textbf{0.7162}  & 0.1838 & 0.2252 & 0.3065 & 0.2333 & \textbf{0.3454}  \\
Credit Approval Dataset & 0.6670 & 0.6584 & 0.6068 & 0.6546 & \textbf{0.7034}  & 0.2869 & 0.2449 & 0.2361 & 0.1816 & \textbf{0.3389}  \\
German Credit Dataset & 0.5045 & 0.5116 & 0.5598 & 0.5741 & \textbf{0.5501}  & 0.0038 & 0.0125 & 0.0063 & 0.0025 & \textbf{0.0218}  \\
Adult Dataset & 0.6037 & 0.6261 & 0.6247 & 0.6258 & \textbf{0.6202}  & 0.0921 & 0.0012 & 0.0054 & 0.0038 & \textbf{0.0920}  \\
\hline
\end{tabular}
\end{table*}

 We vary the choices of the $\kappa_1,\kappa_2$ and $L$ to obtain the feature maps for the different datasets. The feature maps are used to derive clusters using the simple K-means clustering.  Comparing the RI and NMI  for each of the choices we decide on the optimal value of the hyper-parameters. The computed value of the RI and NMI as a result of these selections are given in table \ref{selection_of_hyperparameters}.
The values of the binary cross entropy loss using the soft-max values and mean square error loss function for the different epochs are calculated. The estimated loss functions for the two mentioned deep hidden layer networks are give for the 4 datasets in Fig \ref{loss}  for both training loss and validation loss.
\subsection{Clustering Mixed Type of Data}
We judge the quality of unsupervised feature maps based on their performance in clustering the data. 
\subsubsection{Clustering Results}
  The  latent feature maps are obtained from the two deep encoder decoder networks using the optimal choices of hyper parameters given in Table \ref{selection_of_hyperparameters} . 
  The dimension of the final feature map  obtained from the locality preserving transformation must be carefully controlled. We vary the number of dimensions of the final feature map by selecting different number of eigen vectors in \eqref{gen} and report the clustering results in terms of RI and NMI in Table \ref{vary_ev}. The optimum choice of the feature dimension is shown in bold font. It is clearly observable that both the metrics, RI and NMI initially improve with increasing  dimension of the feature maps and then declines later. Initial choices of feature dimensions are contributing to important information. The noise accumulated due to additional feature dimensions, later over-compensates the separability of the data. As a result, the neighbourhood information gets increasingly blur with higher dimensions. Moreover, the simple K-means clustering method being heavily relying on squared distance between data points is not quite robust to high dimensions of input features. Hence it is important to control the dimension of the feature map for optimal clustering.
  \par
  These optimal choices of the dimensions of the feature map are taken for clustering the datasets with varying number of clusters. For the 4 datasets, varying the number of clusters, the corresponding RI and NMI are reported in Table \ref{vary_clusters}. Values of RI are shown to be decreasing with increase in number of clusters, whereas the values of NMI is increasing. Typically, RI tends to show higher values for equal sized big clusters. On the contrary, NMI shows higher values for  unequal sized clusters, especially when there are multiple smaller clusters (see \cite{ri_nmi}). If the number of clusters are increased, there is chance of having multiple smaller clusters and as a result NMI is tending to higher values.  
\subsubsection{Previous Work on Clustering Mixed Data}
We compare the quality of the derived clusters using the proposed feature maps with some of the competitor clustering techniques for mixed data types available in the literature.
Some of the early approaches such as \cite{sbac} proposed similarity based agglomerative
clustering which performs well for mixed type of
data.  A class of methods  developed similar to {\it K-means} such as {\it K-Prototype} algorithm, aimed at mixed data types, combines the K-Means and K-Mediod clustering methods (see.\cite{proto}). 

The more recent methods target at finding  subspace of the variables where the true clusters are residing. \cite{wkm} suggested a modification in the traditional {\it K-means} by updating weights of the variables based on current data partition in the iterations. We denote this weighted version of {\it K-means} clustering as {\it WKM}.
\cite{ewkm} extends the {\it WKM} clustering further by calculating weight of each variables in each cluster separately by including the weight entropy in the objective function. We denote this strategy of clustering as {\it EWKM}. The {\it OCIL} algorithm  is based on object-cluster similarities. The similarities of the objects in each of the clusters are computed separately for numerical and categorical variables (see \cite{ocil}). In order to incorporate the varying contribution of different variables in forming the clusters, {\it OCIL} is improved by \cite{wocil}. In this approach, the weights of each of the variables for each cluster are dynamically updated during the learning epochs to optimize the object-cluster similarity. We denoted this strategy by {\it WOCIL}.
\par
\subsubsection{Performance Comparison of Clustering} For mixed type of data points, subspace selection or suitably combining different similarity metrics between the points improve the clustering significantly. However,  the present work is more focused on deriving high quality dense feature map of mixed type of data. If the feature maps are of good quality, it is expected to positively impact clustering results even for simple {\it K-Means} clustering. We continue to give more importance to the quality of the proposed feature map and choose simple K-means algorithm on the feature maps for clustering the data points. 
We denote the  algorithm using the simple K-means clustering on the proposed feature map as {\it KMFM}. In the comparison study, for each of the 4 datasets, we run {\it KMFM} along with the sophisticated competitor clustering algorithms: {\it WKM, EWKM, OCIL, WOCIL}. We report the results in terms of RI and NMI in Table \ref{comparison}. 
We choose the feature dimensions and values of the hyper parameters of the network to be the optimized values as listed in Table \ref{selection_of_hyperparameters} for deriving the feature maps. 
\par
At a high level, the proposed method {\it KMFM} always outperforms the competitors with respect to NMI. The performance of {\it KMFM} is, in most cases, neck to neck, if not better in comparison to the competitors with respect to RI. For {\it Heart Disease Dataset} and {\it Credit Approval Dataset} the proposed {\it KMFM} outperforms all the competitor clustering techniques with respect to RI and NMI both. The second best method with respect to RI turns out to be {\it WOCIL} for {\it Heart Disease Dataset}, where as it is {\it WKM} for {\it Credit Approval Dataset}. Now if we turn to the index NMI, it is {\it OCIL} which turns out to be the second best method for {\it Heart Disease Dataset} and for {\it Credit Approval Dataset} it is {\it WKM}. For the {\it German Credit Dataset}, with respect to RI, the proposed {\it KMFM} outperforms {\it WKM} and {\it EWKM}, although it is slightly underperforming in comparison to {\it OCIL} and {\it WOCIL}. However, with respect to NMI, the proposed {\it KMFM} significantly outperforms all the competitors. When we analyse the results for the {\it Adult Dataset}, the proposed method still manages to outperform all the competitors with respect to NMI. However, except for {\it WKM}, the proposed {\it KMFM} is slightly underperforming at third decimal point in comparison to the other three competitors {\it EWKM, OCIL} and {\it WOCIL} with respect to the index RI.
\par
Now if we focus on only the 4 competitor methods, {\it WOCIL} has almost always outperformed among themselves with respect to RI.  This clearly indicates that balanced set of clusters indeed reside in  different subspaces of the set of variables. However, if we focus on the {\it Adult Dataset} the performance  among the different clustering algorithms is minimal indicating that the dataset may not have subspace clusters. In all the cases, the proposed method {\it KMFM} either outperforms or almost neck to neck with  the competitors with respect to RI.The performance metric RI typically tends to show higher values for balanced set of big cluster.  This is an indication to the fact that balanced set of clusters are efficiently identifiable  through more involving feature representation of the data points and a simple K-means clustering as compared to more sophisticated subspace or weighted clustering techniques.
\par
Considering NMI, which tends to show higher values for unbalanced clusters with many smaller clusters being present, the competitors {\it WKM, EWKM, OCIL, WOCIL} have no clear winner. However, the proposed method {\it KMFM} comprehensively outperforms  all of them in most cases except for {\it WKM} in {\it Adult Dataset}.
The result clearly indicates that nonlinear combination of mixed space variables can discover unequal sized unbalanced clusters quite consistently as compared to sophisticated competitor subspace clustering techniques.
\section{Discussion}
Mixed type of variables often give rise to the challenge of combining the information available in the mixed data product space to get good feature maps.  Important information often relates to higher order properties like interactions between the variables and nonlinear mutual dependencies. There is some information available in the one type of variables that correlates with other type. With this hypothesis in mind, we have proposed two deep encoder decoder networks keeping categorical and numerical variables at the two ends of the networks in turn. The information available in one type of variables that connects or correlates with the other type should get encoded in the intermediate hidden layers of the networks. The hidden layers connect the two types of variables through various non-linear transformations of the features to give latent feature maps.  With these estimated latent features we proposed a linear  transformation on them to make sure that the local geometry of the actual data points are also consistent in the final feature maps. These feature maps can be further used for various statistical learning problem. When it comes to clustering the data points, there are various clustering algorithms which can work well. But how these clustering algorithms can produce feature maps combining the information from the mixed space is not so obvious. In this work we focus on getting such high quality dense feature maps for mixed space data. Interestingly, the numerical comparison study shows that the proposed feature map is equally efficient, if not more efficient, in producing similar quality clusters with simple K-means algorithm as produced by the more sophisticated mixed space clustering algorithms.  This finding leads to a very interesting  direction of research in general. One can imagine that with a suitable architecture of representation learning, informative and dense feature maps are discoverable without additional supervision. A simple learning algorithm like { K-means} can take such high quality feature maps as input to produce good quality results. The results, in this case, the clusters, can be  comparable to other complex sophisticated strategies directly which take raw data points as input.

\bibliographystyle{IEEEtran}
\bibliography{biblio_mixed}

\end{document}